# A Machine Learning Framework for Register Placement Optimization in Digital Circuit Design

Karthik Airani, Rohit Guttal
Department of Silicon Engineering, Micron Technology
8000 South Federal Way, Boise, Idaho

## Abstract
In modern digital circuit back-end design, designers heavily rely on electronic-design-automoation (EDA) tool to close timing. However, the heuristic algorithms used in the place and route tool usually does not result in optimal solution. Thus, significant design effort is used to tune parameters or provide user constraints or guidelines to improve the tool performance. In this paper, we targeted at those optimization space left behind by the EDA tools and propose a machine learning framework that helps to define what are the guidelines and constraints for registers placement, which can yield better performance and quality for back-end design. In other words, the framework is trying to learn what are the flaws of the existing EDA tools and tries to optimize it by providing additional information. We discuss what is the proper input feature vector to be extracted, and what is metric to be used for reference output. We also develop a scheme to generate perturbed training samples using existing design based on Gaussian randomization. By applying our methodology, we are able to improve the design runtime by up to 36% and timing quality by up to 23%.

## 1. Introduction
Physical design in very-large-scale-integrated (VLSI) circuit has been increasingly difficult, due to technology scaling, and more restrictive place and route rules. Current electronic-design-automation (EDA) tools either run very slow or could not achieve satisfactory results (e.g. timing is broken with huge number of total negative slack (TNS), etc.). Conventional heuristic algorithms do not scale well with the technology advance and design complexity increase. And because of the limitation of the existing algorithms, no significant break through was observed over the past years.

Recently, machine learning has become a hot topic in various research fields, such as natural language processing, image recognition, etc. [1-2]. We found VLSI physical design is another aspect of area where machine learning can potentially leverage its best advantages in applications. And because of the nature of machine learning algorithms, they scale well with design complexity. An interesting observation is that: according to Moore's Law [3], the number of transistors in a dense integrated circuit doubles approximately every two years, therefore if machine learning technique is applied during VLSI design, the size of data training smaple also doubles every two years. Moreover, Moore's Law can also be interpreted as the computational power doubles every two years. Since for most machine learning algorithms, the training time is linearly proportional to the sample size, thus eventually, the training time remains a constant, assuming no algorithm runtime complexity improvement. With this observation, we see machine learning to be a highly promising technique that can constantly evolve the design with technology change.

In this paper, we mainly focus on the place stage of integrated circuit physical design, and discuss how to apply machine learning technique to improve both the performance and runtime of physical placement. The key is to select proper set of input feature vector that is highly correlated to the outcome of the placement such that the learning algorithm can update efficiently. Also the training model selection is critical to yield better performance of the placement result. The better placement result also helps later optimizations such as electro migration improvement [4].

In Seciont 2, we discuss input feature selection and extraction, which uses a simplified logic chain to represent the physical and logical property of a register. In Section 3, we discuss the target output extraction and propose a Gaussian randomization scheme for generating meaningful training data within feasible turn around time. In Section 4, we briefly compared different machine learning models and their performance combined with EDA tools. In Section 5, we presented experimental data and we concluded in Section 6.

## 2. Input Feature Selection

In typical computer architecture with pipeline, registers are usually the reference points for the data path. Date is launched from one register (a.k.a. starting point) and captured by another register (a.k.a. end point), various logical operations happen along the data path, also referred to as a timing path. Once the registers are placed, logical standard cells can be easily placed around the registers to close circuit timing. Also, the registers remain the same after synthesis and throughout place and route stages, while logical cells can be optimized or buffered that lose track in the final database. Therefore, we focus on feature extraction on those registers and use them as reference point for placement optimization. The other benefits is that by excluding the regular logical cells, we reduce the training sample size significantly, keeping only the highly relevant ones.

The preceding and following logic of a register can be used to fully describe its logical feature. As shown in Figure 1, register to register logic is the only objects seen by a register.

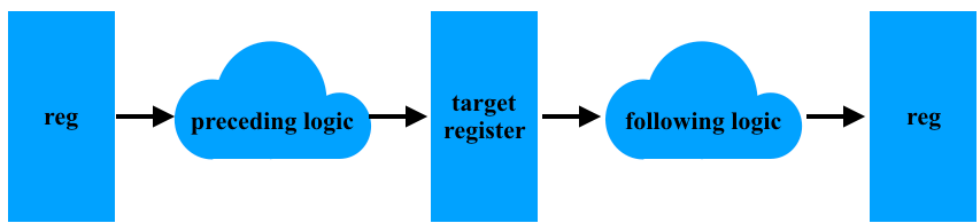

**Figure 1. Register to register logic**

However, the physical information is completely missing in the above scenario. To extract physical feature of a register, we need to trace back to an input port and trace forward to an output port of the design module. Here we assume port locations are given. Although co-optimization of port location and register placement is possible, it is a separate design optimization technique and is beyond the scope of this paper. Therefore the complete logical chain can be used to describe the combined physical and logical feature of a register, which is illustrated in Figure 2.

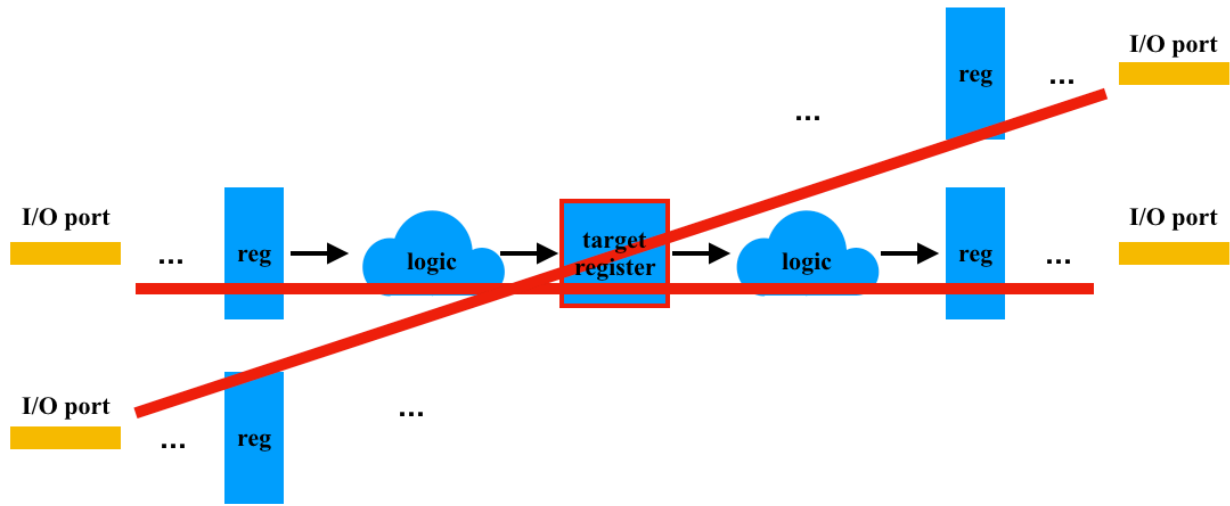

Figure 2. Complete logic chains

However, there are two limitations of the above input feature set: (1) there could be millions of elements along a single logical chain; (2) a register can belong to multiple logical chains, and one may need to recursively trace every branch, which can lead to trace through the whole design. To overcome the above restrictions and derive a feasible input feature vector, we apply two simplifications: (1) the total logical depth is used to simplify the representation of the elements along the logical chain; (2) we select the top 100 deepest logical chains for each register and filter out the remaining ones. Here, the logic depth is defined as in a digital circuit the maximum number of basic gates (AND, OR, INV, etc.) a signal need to travel from source register to destination register [5]. Therefore, logic depth is proportional to the delay, as the delay of basic gates can be approximated as constant. Also, it can be that the deepest logical chain represents the most critical paths. With these two simplifications, the input feature vector of registers can be efficiently extracted and trained within reasonable amount of runtime.

## 3. Regression Target Extraction

Once input feature vector is extracted for each register, we need to decide what could potentially be a good metric as the target value for output. A naive metric would be the x and y coordinates for each register given a timing-closed finalized design. However, this would require run through place and route for various designs to derive a meaningful number of reference outputs. For today's typical digital circuit design, a full place and route run will take 5~10 days to finish. To overcome the turn around time issue, we propose a new metric that adds the worst slack as one additional input feature and still keep x and y coordinates as the output reference metric. In this way, we are able to generate training samples easily by perturbing the design using Gaussian randomization [6-7] (e.g. manually move some registers and followed re-legalize and ECO route). Another advantage is that during the prediction phase, we can always set the worst slack to 0 to constrain the design for timing-closure. Figure 3 shows how to generate a new training sample based on an existing timing-closed design.

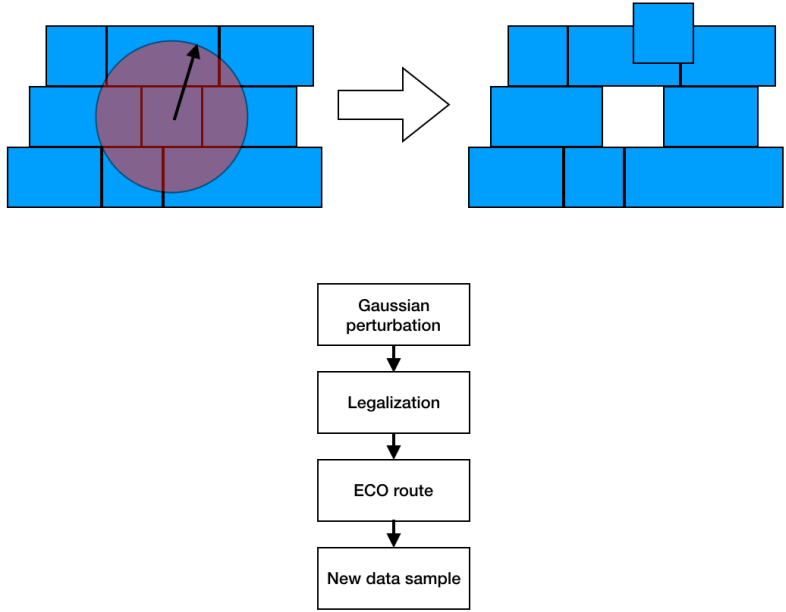

**Figure 3. Gaussian perturbation for deriving new data sample**

Based on above analysis, we formally define the input feature vector and output target vector to be:

*Logic chain := {(x,y) of input port, logic depth, (x,y) of output port}*

*Input vector := {top 100 logic chains, worst slack}*
*Output vector := {(x,y) of register location}*

## 4. Training Model Selection

There exists various machine learning models, but no detailed theoretical analysis nor comprehensive comparisons were provided by the research community. However, we can roughly divide the major machine learning models into a few categories depending on the algorithms used underneath. In our experiments, we pick three typical machine learning models: kernel ridge regressor, support vector regressor and random forest regressor. Kernel ridge regression (KRR) [8] combines Ridge Regression (linear least squares with l2-norm regularization) with the kernel trick. It thus learns a linear function in the space induced by the respective kernel and the data. For non-linear kernels, this corresponds to a non-linear function in the original space. Support vector regression (SVR) [9] are supervised learning models with associated learning algorithms that analyze data used for regression analysis. Given a set of training examples, each marked as belonging to one or the other of two categories, an SVR training algorithm builds a model that assigns new examples to one category or the other, making it a non-probabilistic regressor. A random forest regression (RFR) [10] is a meta estimator that fits a number of classifying decision trees on various sub-samples of the dataset and use averaging to improve the predictive accuracy and control over-fitting. The sub-sample size is always the same as the original input sample size but the samples are drawn with replacement. Prediction accuracy, error and runtime are compared across the three regressors. Figure 4 shows the comparison results.

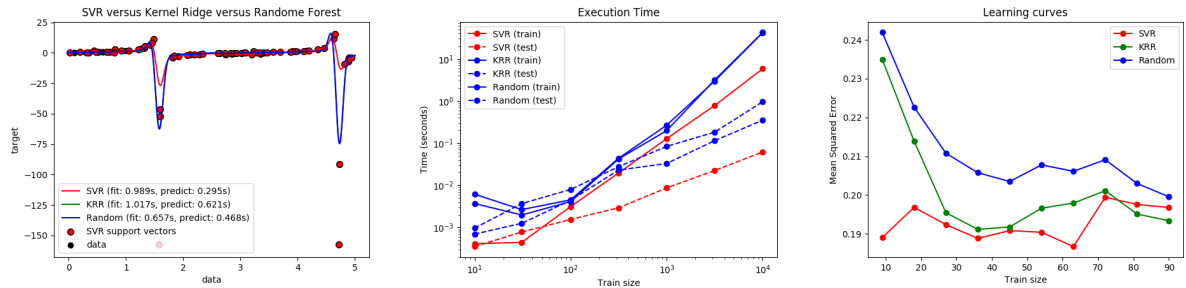

**Figure 4. Machine Learning Model Comparisons**

We can observe that KRR and RFR have similar fitting curves, where they can handle outliers better compared to SVR. However, SVR generally has a better runtime among the all. Also, RFR shows a smoother learning curve, while both KRR and SVR are seeing overfitting issues. Note, the comparison is purely for the regressor prediction accuracy, while the final merit of the regressor needs to be evaluated by the timing quality of reference (QoR). Therefore, there is one level of conversion from the regressor to place and route tool. This conversion is uncontrollable and irrelevant to theoretical analysis and can only be evaluated by experimental results. In our experiments, we found that the RFR has the best performance in practice and therefore the experimental results we presented in Section 5 are based on RFR + regular physical design using EDA tools.

## 5. Experimental Results

We used one block from the open source design or1200_fpu_arith [11] as our benchmark testcases. The logical synthesis is done using Synopsis Design Compiler [12] and physical synthesis is done using Synopsis ICC2 [13]. Sign-off extraction is done using Synopsis Star-RC [14] and timing analysis is done through PrimeTime [15]. We use Gaussian randomization to generate perturbed training samples. Each time, we randomly pick registers to move its placement, while the movement is determined by Gaussian distribution. To generate more realistic training samples, after each perturbation, we always refine placement and run ECO route to derive physical clean design and rerun PrimeTime analysis to calibrate its timing QoR.

Once machine learning model is trained and calibrated, we follow the below steps to apply to a new design:
(1) Run machine learning prediction to derive ideal location of each register;
(2) Use the predicted location as seed placement, extend each (x,y) coordinate to grow a 2um by 2um soft bound for each register;
(3) Place the whole design using the soft bound guide followed by regular place and route operations.

The machine learning runs are compared against baseline run using another testing block or1200_alu. The comparison metric includes runtime and timing QoR. Comparison summary is provided in Table 1. It can be observed that run time for place stage has improved a lot, also the runtime for post place stage (e.g. clock and route) improved. This is because the seed placement of machine learning predicted result also gives better overall routing congestion map and less hot spot area. As expected, timing QoR improved significantly for both total negative slack (TNS) and worst negative slack (WNS). One thing to note is that total power also reduced slightly, this is because the machine learning predicted results gives an overall better physical design starting point, where buffer count/sizes and routing length can be reduced that helps to bring down the total power of the circuit block.

Table 1. Runtime and timing QoR comparisons

| or1200_alu | Baseline run | Machine learning run |
|---|---|---|
| Runtime (place) | 2d 10hrs 32mins | 20hr 12mins |
| Runtime (post-place to finish) | 5d 11hrs 24mins | 4d 5h 43mins |
| Total negative slack | -73.859ns | -57.069ns |
| Worst negative slack | -128ps | -103ps |
| Total power | 1.784mW | 1.669mW |

## 6. Conclusion

In this paper, we propose a machine learning framework that helps improve digital circuit register placement. The register logic chains, depth and path slack is extracted as input feature vector. The eventual (x,y) coordinate of a register is considered as target outputs. The proposed approach simplifies the physical features of a register and extract only major contributors. To obtain reasonable training sample sizes, we also develop a Gaussian randomization based algorithm to perturb the design for efficiently generating more training data. Comparisons among different machine learning models are provided and summarized. We also provide the steps to best utilize our proposed framework for designers to improve their place and route run of a digital block. The results show that runtime, timing QoR and total power are all improved using the machine learning framework to help physical design. We also encourage researchers to look into wider and deeper machine learning applications in digital circuit design methodology/algorithms to close the gap of optimization space left by EDA tools.